\documentclass[letterpaper]{article} % DO NOT CHANGE THIS
\usepackage{aaai2026}  % DO NOT CHANGE THIS
\usepackage{times}  % DO NOT CHANGE THIS
\usepackage{helvet}  % DO NOT CHANGE THIS
\usepackage{courier}  % DO NOT CHANGE THIS
\usepackage[hyphens]{url}  % DO NOT CHANGE THIS
\usepackage{graphicx} % DO NOT CHANGE THIS
\usepackage{natbib}  % DO NOT CHANGE THIS AND DO NOT ADD ANY OPTIONS TO IT
\usepackage{caption} % DO NOT CHANGE THIS AND DO NOT ADD ANY OPTIONS TO IT

\usepackage{algorithm}
\usepackage{algorithmic}
\usepackage{booktabs} 
\usepackage{multirow}
\usepackage{amssymb}
\usepackage{amsthm}
\usepackage{amsmath}
\usepackage{newfloat}
\usepackage{listings}
\usepackage{cleveref}
\DeclareCaptionStyle{ruled}{labelfont=normalfont,labelsep=colon,strut=off} % DO NOT CHANGE THIS
\floatstyle{ruled}
\newfloat{listing}{tb}{lst}{}
\floatname{listing}{Listing}
\usepackage{color} %(if used in text)

\title{Cross-modal Proxy Evolving for OOD Detection with Vision-Language Models}
\author{
   Hao Tang\textsuperscript{\rm 1},
   Yu Liu\textsuperscript{\rm 1},
   Shuanglin Yan\textsuperscript{\rm 2}\thanks{Corresponding author.},
   Fei Shen\textsuperscript{\rm 3},
   Shengfeng He\textsuperscript{\rm 4},
   Jing Qin\textsuperscript{\rm 1}
}
\affiliations{
    \textsuperscript{\rm 1}~Centre for Smart Health, The Hong Kong Polytechnic University\\
    \textsuperscript{\rm 2}~College of Information Science and Technology, Nanjing Forestry University\\
    \textsuperscript{\rm 3}~NExT++ Research Center, National University of Singapore\\
    \textsuperscript{\rm 4}~School of Computing and Information Systems, Singapore Management University
}

\usepackage{bibentry}
\begin{document}
\makeatletter
\def\copyright@on{}
\makeatother

\maketitle

\begin{abstract}
Reliable zero-shot detection of out-of-distribution (OOD) inputs is critical for deploying vision-language models in open-world settings. However, the lack of labeled negatives in zero-shot OOD detection necessitates proxy signals that remain effective under distribution shift. Existing negative-label methods rely on a fixed set of textual proxies, which (i) sparsely sample the semantic space beyond in-distribution (ID) classes and (ii) remain static while only visual features drift, leading to cross-modal misalignment and unstable predictions. In this paper, we propose \textbf{\texttt{CoEvo}}, a training- and annotation-free test-time framework that performs bidirectional, sample-conditioned adaptation of both textual and visual proxies. Specifically, \texttt{CoEvo} introduces a \textit{proxy-aligned co-evolution} mechanism to maintain two evolving proxy caches, which dynamically mines contextual textual negatives guided by test images and iteratively refines visual proxies, progressively realigning cross-modal similarities and enlarging local OOD margins. Finally, we dynamically re-weight the contributions of dual-modal proxies to obtain a calibrated OOD score that is robust to distribution shift. Extensive experiments on standard benchmarks demonstrate that \texttt{CoEvo} achieves state-of-the-art performance, improving AUROC by $1.33\%$ and reducing FPR95 by $45.98\%$ on ImageNet-1K compared to strong negative-label baselines.
\end{abstract}

\begin{links}
    \link{Code}{https://github.com/yuleoliu/CoEvo}
\end{links}

\section{Introduction}

Machine learning systems deployed in real-world environments frequently encounter inputs from previously unseen classes, commonly referred to as \textit{out-of-distribution} (OOD) data. These inputs often differ significantly from the pre-defined \textit{in-distribution} (ID) categories observed during training. When presented with such data, models tend to produce overconfident yet incorrect predictions~\cite{scheirer2012toward,nguyen2015deep}, posing substantial safety and reliability concerns in high-stakes applications such as healthcare and autonomous driving. OOD detection aims to mitigate these risks by identifying and rejecting OOD inputs, thereby enhancing the robustness of downstream decision-making systems.

\begin{figure}[t!]
\begin{center}
% left bottom right top
\includegraphics[width=0.45\textwidth]{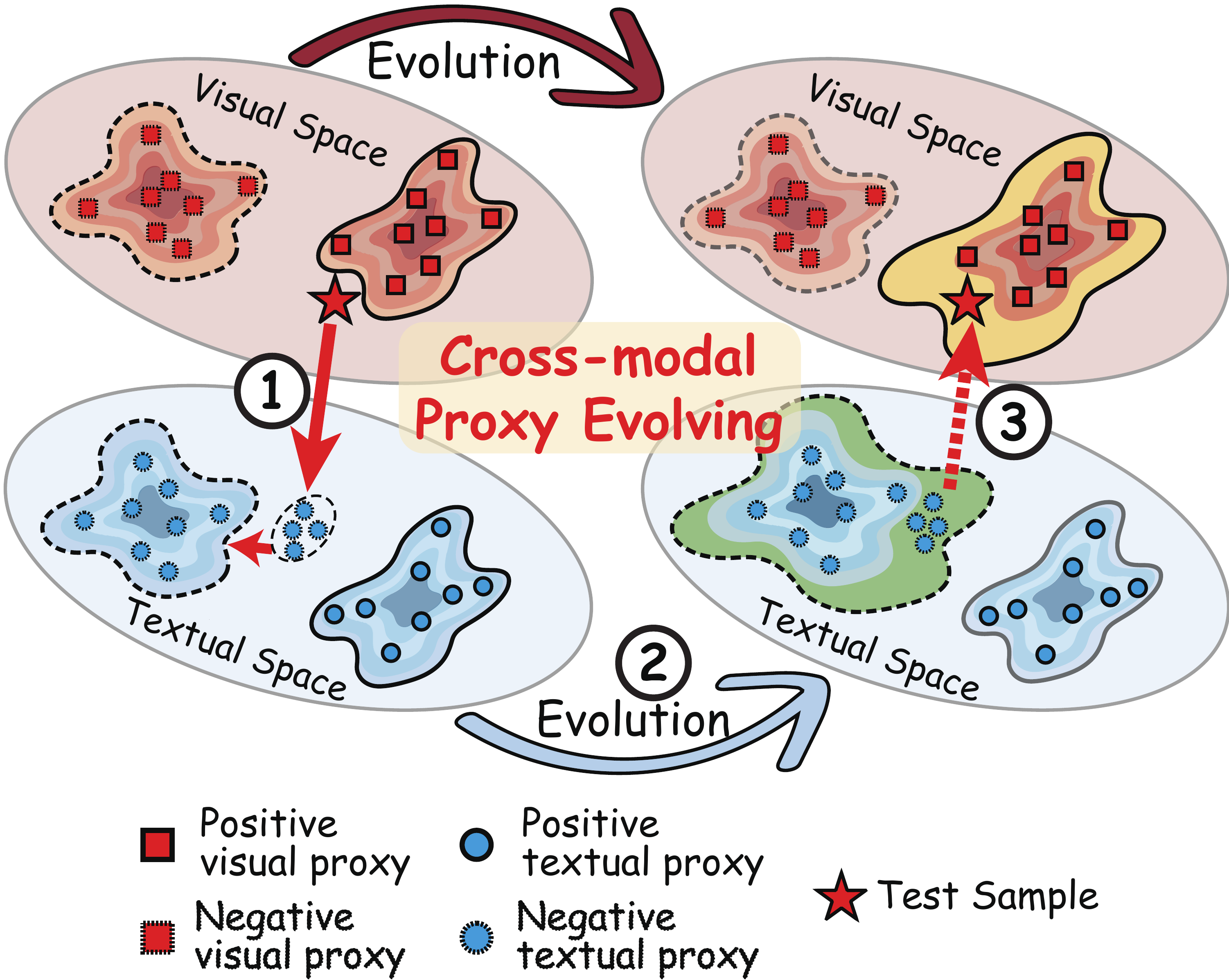}
\end{center}
  \caption{
\textbf{Proxy-aligned co-evolution.} For each test sample, textual negatives are dynamically mined to expand the occupied space around its semantic context, while visual positive/negative proxies are updated online. This joint evolution maintains aligned cross-modal similarities under distribution shift, enabling robust zero-shot OOD decisions.
 }
\label{fig:evolution}
\end{figure}

Traditional vision-based OOD detection methods primarily operate within the visual domain, relying solely on image features while overlooking the rich semantic information embedded in class labels~\cite{hendrycks2016baseline,lee2018simple,liang2017enhancing}. Recent advances in vision-language pre-training, particularly CLIP~\cite{radford2021learning}, have enabled a multi-modal paradigm that leverages both visual and textual information in zero-shot OOD detection~\cite{esmaeilpour2022zero,ming2022delving,wang2023clipn}. Within this paradigm, a prominent line of work, known as \emph{negative-label} methods, constructs a pool of textual labels that are semantically dissimilar from ID classes and employs them as textual proxies for ``not-ID'' concepts. For example, NegLabel~\cite{jiang2024negative} selects negatives from WordNet~\cite{fellbaum1998wordnet}, while CSP~\cite{Chen0X24} augments labels with descriptive adjectives to synthesize semantically unrelated superclasses. A test image is then classified as OOD if its similarity to these negative labels surpasses that to ID labels.

While negative-label approaches have shown promise, their static design introduces two fundamental limitations: 
\textbf{(i) Unmodeled negative space:} A globally fixed negative set sparsely samples the vast semantic space beyond ID classes, leaving many informative, sample-specific negatives unrepresented during inference.  
\textbf{(ii) Modality misalignment:} Under test-time distribution shift, visual features shift to the new domain, whereas textual negatives remain fixed to presetting priors. This desynchronization distorts the cross-modal similarity geometry and destabilizes decision thresholds. 
Recent work such as AdaNeg~\cite{ZhangZ24} partially addresses the first issue by constructing visual proxies using encountered test samples and jointly leveraging visual-textual evidence during scoring. However, it still employs \emph{fixed textual negatives}: the negative words are preselected (e.g., far from ID labels in the text space) and kept fixed during inference. That is, adaptation is inherently \emph{one-sided}: visual proxies adapt to test data, whereas textual negatives remain static. Consequently, cross-modal geometry is only partially realigned, and a substantial portion of the negative space remains unmodeled.

We argue that robust zero-shot OOD detection requires \emph{bidirectional, sample-conditioned adaptation} of both modalities. Specifically, textual negatives should dynamically adapt to the current test sample and domain, while visual proxies should expose OOD structure as data arrive,without updating the backbone parameters or relying on labeled OOD samples. This motivates our proposed \textit{proxy-aligned co-evolution} mechanism (Fig.~\ref{fig:evolution}),  where visual cues guide the mining of contextual textual negatives, and the updated textual proxies, in turn, refine the visual decision boundary in a closed-loop manner.

To instantiate this principle, we introduce \texttt{CoEvo} (\emph{Cross-modal Proxy Evolving}), a test-time zero-shot OOD detection framework illustrated in Fig.~\ref{fig:overview}. \texttt{CoEvo} maintains two modality-specific proxy caches (textual and visual), each organized as positive/negative queues  that are updated online through the proposed \textit{proxy-aligned co-evolution} mechanism. These updates form a closed-loop process that iteratively aligns cross-modal similarities and enlarges local OOD margins. Additionally, we adaptively re-weight contributions from the dual-modal proxies to produce a calibrated OOD score. Extensive experiments on standard benchmarks validate the effectiveness of \texttt{CoEvo}. Notably, on the large-scale ImageNet dataset, \texttt{CoEvo} achieves a $1.33\%$ improvement in AUROC and a $45.98\%$ reduction in FPR95 over the best-performing negative label-based baselines.

We summarize our contribution as follows:
\begin{itemize}
    \item We propose \texttt{CoEvo}, a zero-shot OOD detection framework that constructs semantically aligned ID/OOD proxy caches at test time by jointly leveraging visual and textual modalities.
    \item We introduce a proxy-aligned co-evolution mechanism  that performs sample-conditioned, bidirectional adaptation of modality-specific proxies, mitigating cross-modal misalignment under distribution shift.
    \item Extensive experiments on widely used large-scale benchmarks show state-of-the-art performance; e.g., on ImageNet, \texttt{CoEvo} improves AUROC by $1.33\%$ and reduces FPR95 by $45.98\%$ over strong negative-label baselines.
\end{itemize}

\section{Related Work}

\noindent \textbf{OOD Detection with Visual Modality.}
Existing visual OOD detection approaches can be broadly classified into three categories: score-based~\cite{huang2021mos, wang2022vim, sun2021react}, distance-based~\cite{tack2020csi, du2022siren, ming2022exploit}, and generative-based methods~\cite{ryu2018out, kong2021opengan}.
Among them, score-based methods are particularly prominent, as they introduce various scoring mechanisms to effectively discriminate between ID and OOD samples. Representative scoring strategies include confidence-based~\cite{sun2021react}, discriminator-based~\cite{kong2021opengan}, energy-based~\cite{liu2020energy, wang2021energy}, and gradient-based scores~\cite{huang2021importance}. In contrast, distance-based methods identify OOD samples by calculating distances between the test sample and the nearest ID sample~\cite{tack2020csi}, or distances to precomputed ID prototypes~\cite{tao2023non}.  Common metrics in this category include KNN~\cite{sun2022out, ming2022exploit} and RBF kernels~\cite{AmersfoortSTG20}. Despite the notable progress achieved, conventional single-modal visual OOD detection methods generally rely on manually labeled ID images and largely overlook the potential benefits derived from textual information integration.

\noindent \textbf{OOD Detection with Dual Modalities.}
Recent zero-shot OOD detection methods increasingly leverage vision-language models (VLMs) to incorporate semantic cues from textual information. Early approaches such as ZOC~\cite{esmaeilpour2022zero} and CLIPN~\cite{wang2023clipn} employ textual descriptions or auxiliary encoders to detect OOD samples but rely solely on positive in-distribution (ID) labels, often producing overly optimistic similarity scores for unknown classes. To alleviate this limitation, several post-hoc scoring mechanisms have been proposed to enhance OOD discrimination. For instance, MCM~\cite{ming2022delving} computes the maximum softmax score over CLIP similarities, while NegLabel~\cite{jiang2024negative} refines this approach by mining negative samples from external sources. CSP~\cite{Chen0X24} expands the label space with a conjugated semantic pool, and AdaNeg~\cite{ZhangZ24} dynamically adapts negative labels online using encountered OOD samples. However, these approaches construct textual negatives independently of the visual OOD features observed during inference, leading to a persistent \emph{modality misalignment}: negative textual proxies fail to accurately capture the distribution of unseen images, thereby limiting detection robustness. Our method addresses this gap through a cross-modal proxy evolving framework that jointly refines textual and visual proxies, ensuring adaptive and semantically consistent representations for improved zero-shot OOD detection.

\section{Methodology}
	
\begin{figure*}[t!]
\begin{center}
\includegraphics[width=0.95\textwidth]{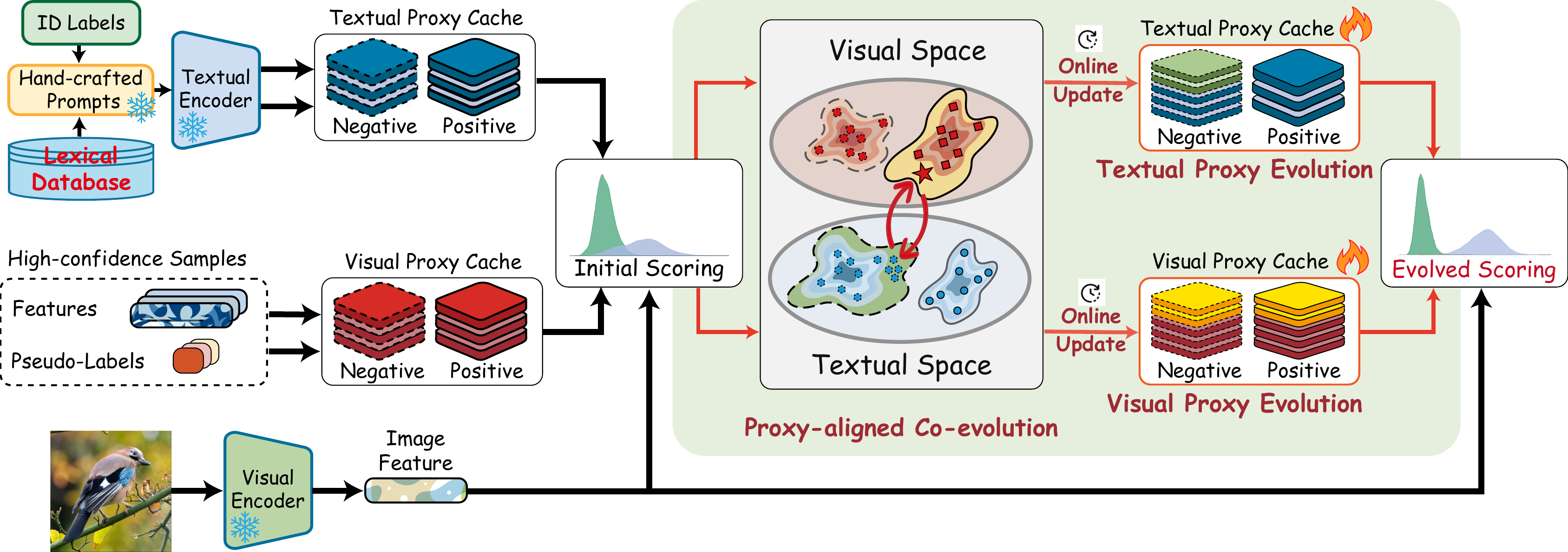}
\end{center}
\vspace{-0.1cm}
  \caption{Pipeline of the proposed cross-modal proxy co-evolving framework (\texttt{CoEvo}). It dynamically updates both visual and textual proxy caches based on high-confidence samples, enabling bidirectional alignment and robust zero-shot OOD detection.}
% \vspace{-0.2cm}
\label{fig:overview}
\end{figure*}

\paragraph{Problem Formulation.}~\label{sec3.1}
Let $\mathcal{Y}_{\mathrm{ID}} = \{y_1, \dots, y_K\}$ denote the label set of $K$ ID classes, and let $x \in \mathcal{X}$ represent an input image. Zero-shot OOD detection with vision–language models~\cite{ming2022delving,wang2023clipn,jiang2024negative}, such as CLIP~\cite{radford2021learning}, aims to determine whether $x$ belongs to any class in $\mathcal{Y}_{\mathrm{ID}}$ or originates from an unseen class (OOD), without requiring any training samples or prompt tuning. The core objective is to design an ID confidence function $\mathcal{S}: \mathcal{X} \to \mathbb{R}$ that assigns higher values to ID samples than to OOD ones. Given a threshold $\theta \in \mathbb{R}$, the detector predicts \textit{ID} if $S(\mathbf{x}) \ge \theta$ and \textit{OOD} otherwise.

\subsection{Overview}
Existing negative label-based methods typically rely on \emph{static textual proxies}, i.e.,~a fixed set of textual embeddings that serve as negative semantic anchors. However, these static proxies leave portions of the negative semantic region uncovered, i.e., an \emph{unmodeled negative space}.
Furthermore, due to these proxies remain unchanged at test time, they cannot adapt to image features that drift under distribution shift, which
induces \emph{modality misalignment} between image and text similarities. To address these limitations, we introduce \textit{Cross-modal Proxy Evolving} (\texttt{CoEvo}), a test-time framework that constructs OOD proxies by jointly exploiting visual and textual modalities, as illustrated in Fig.~\ref{fig:overview}. \texttt{CoEvo} maintains two online caches: a \emph{textual proxy cache} that stores positive/negative textual proxies, and a \emph{visual proxy cache} that stores positive/negative visual proxies. A \emph{proxy-aligned co-evolution} mechanism then couples the two caches so that images teach text where the non-ID regions lie, and the updated text in turn regularizes visual decisions.

\subsection{Textual Proxy Cache}

Textual proxies offer a powerful means of encoding semantic priors for zero-shot OOD detection. Leveraging the expressive capability of pre-trained vision-language models (e.g., CLIP), we construct two complementary textual proxy queues: a \emph{positive} proxy queue for ID classes, and a \emph{negative} proxy queue for OOD concepts.

\paragraph{Positive Proxy Queue.}  
Let the ID label set be defined as $\mathcal{Y}_{\mathrm{ID}} = \{y_1, \dots, y_K\}$, where each $y_k$ denotes a semantic category (e.g., \textit{cat}, \textit{dog}, \textit{bird}), and $K$ denotes the number of ID classes. Each class name $y_k$ is converted into a prompt-based textual embedding using a pre-trained CLIP text encoder as $\mathbf{f}_{\mathrm{t},k} = \mathcal{E}_{\mathrm{t}}(\mathcal{T}(y_k)) \in \mathbb{R}^D$, where $\mathcal{T}(\cdot)$ denotes the prompt template (e.g., ``a photo of a \texttt{<class>}''), and $D$ is the embedding dimension. We organize these embeddings into a fixed queue $\mathbf{T}_p \in \mathbb{R}^{K \times D}$, i.e., $\mathbf{T}_p[k] = \mathbf{f}_{\mathrm{t},k}$. The static nature of $\mathbf{T}_p$ reflects the stable semantics of known categories.

\paragraph{Negative Proxy Queue.} 
To model unknown OOD semantics, following NegLabel~\cite{jiang2024negative}, we initialize a negative textual proxy queue $\mathbf{T}_n \in \mathbb{R}^{M \times D}$ with $M$ negative class embeddings sampled from a large-scale lexical corpus $\mathcal{D}$. Specifically, we define a disjoint set of negative labels $\mathcal{Y}_{\mathrm{neg}} = \{\tilde{y}_1, \dots, \tilde{y}_M\}$, where $\mathcal{Y}_{\mathrm{neg}} \cap \mathcal{Y}_{\mathrm{ID}} = \varnothing$, and obtain $\mathbf{T}_n = \mathcal{E}_{\mathrm{t}}(\mathcal{T}(\mathcal{Y}_{\mathrm{neg}})) \in \mathbb{R}^{M \times D}$. Unlike $\mathbf{T}_p$, the negative proxy cache $\mathbf{T}_n$ evolves during inference: new embeddings are dynamically incorporated based on test-time visual observations through our proposed proxy co-evolution mechanism. This adaptivity ensures that $\mathbf{T}_n$ remains responsive to emerging semantic patterns in unseen OOD inputs.

\paragraph{Textual OOD Scoring.}
Given a test image $x$, we obtain its visual embedding via the CLIP image encoder: $\mathbf{f}_{\mathrm{v}} = \mathcal{E}_{\mathrm{v}}(x) \in \mathbb{R}^D$. Assuming that ID inputs are more similar to positive textual proxies and dissimilar to negative ones, we define the textual OOD detection score as:
{\small
\begin{equation}
\label{eq1}
\mathcal{S}_{\mathrm{T}}^{\mathrm{pre}}(x) =
\frac{
\sum_{k=1}^{K} e^{\left( \operatorname{sim}(\mathbf{f}_{\mathrm{v}}, \mathbf{T}_p[k]) / \tau \right)}
}{
\sum_{k=1}^{K} e^{\left( \operatorname{sim}(\mathbf{f}_{\mathrm{v}}, \mathbf{T}_p[k]) / \tau \right)}
+
\sum_{m=1}^{M} e^{\left( \operatorname{sim}(\mathbf{f}_{\mathrm{v}}, \mathbf{T}_n[m]) / \tau \right)}
},
\end{equation}
}
where $\operatorname{sim}(\cdot, \cdot)$ denotes cosine similarity and $\tau$ is a temperature scaling factor. Higher $\mathcal{S}_{\mathrm{T}}^{\mathrm{pre}}(x)$ indicates stronger alignment with known ID semantics, and lower values signal potential OOD samples.

\subsection{Visual Proxy Cache}
While textual proxies provide global semantic anchors, they are inherently limited in representing fine-grained visual variations of unseen data. In particular, zero-shot OOD inputs may exhibit appearance patterns that are poorly described by label embeddings alone. To address this, we introduce \emph{visual proxies} that directly encode instance-level image features and evolve online to complement textual proxies.

\paragraph{Positive Proxy Queue.}
We construct a positive visual proxy queue $\mathbf{V}_p \in \mathbb{R}^{K \times L \times D}$, where $K$ is the number of ID classes and $L$ denotes the number of stored visual instances per ID class. The queue is initially empty, except for the first slot of each class, which is initialized with the corresponding positive textual embedding from $\mathbf{T}_p\in \mathbb{R}^{K \times D}$, i.e., $\mathbf{V}_p[:,0,:] = \mathbf{T}_p$. This provides a semantically meaningful starting point before any labeled images are observed. As inference proceeds, high-confidence ID samples are inserted into the queue, progressively enriching class-specific visual proxies.

\paragraph{Negative Proxy Queue.}
Analogously, we maintain a negative visual proxy queue $\mathbf{V}_n \in \mathbb{R}^{M \times L \times D}$ aligned with the negative textual proxy queue $\mathbf{T}_n \in \mathbb{R}^{M \times D}$. Following AdaNeg~\cite{ZhangZ24}, test samples predicted as OOD with high confidence are enqueued, enabling $\mathbf{V}_n$ to capture the evolving OOD appearance space. A priority queue strategy discards low-similarity or outdated samples, ensuring proxies remain representative of the most informative shifts in the test distribution.

\paragraph{Visual Proxy Aggregation and Scoring.}
Given an input image $x$ with embedding $\mathbf{f}_{\mathrm{v}}$, we compute class-wise aggregated visual proxies via similarity-based attention over the $L$ instances:
\begin{equation}
\label{eq2}
\mathbf{v}^p_k =
\sum_{\ell=1}^{L}
\frac{\exp\big(-\beta \big(1-\mathbf{f}_{\mathrm{v}} \cdot (\mathbf{V}_p^{k,\ell})^\top \big)\big)}
{\sum_{\ell'=1}^{L}\exp\big(-\beta \big(1-\mathbf{f}_{\mathrm{v}} \cdot (\mathbf{V}_p^{k,\ell'})^\top \big)\big)}
\;\mathbf{V}_p^{k,\ell}.
\end{equation}
where $\beta$ controls attention sharpness. Negative proxies $\mathbf{v}^n_m$ are obtained analogously from $\mathbf{V}_n$. The visual OOD score is then defined symmetrically to the textual score:
\begin{equation}
\label{eq3}
\mathcal{S}_{\mathrm{V}}^{\mathrm{pre}}(x) =
\frac{\sum_{k=1}^{K}e^{(\operatorname{sim}(\mathbf{f}_{\mathrm{v}},\mathbf{v}^p_k)/\tau)}}
{\sum_{k=1}^{K}e^{(\operatorname{sim}(\mathbf{f}_{\mathrm{v}},\mathbf{v}^p_k)/\tau)}+
\sum_{m=1}^{M}e^{(\operatorname{sim}(\mathbf{f}_{\mathrm{v}},\mathbf{v}^n_m)/\tau)}}.
\end{equation}
A higher score indicates stronger alignment with ID visual proxies, while lower scores suggest OOD samples.

\paragraph{Multi-modal OOD Score.}
To leverage the complementary strengths of textual and visual modalities, we combine their respective scores into a unified OOD detection score:
\begin{equation}
\label{eq4}
\mathcal{S}_{\mathtt{CoEvo}}^{\mathrm{pre}}(x) = \lambda \mathcal{S}_{\mathrm{T}}^{\mathrm{pre}}(x) + (1 - \lambda) \mathcal{S}_{\mathrm{V}}^{\mathrm{pre}}(x),
\end{equation}
where the hyperparameter $\lambda \in [0.5,1)$ balances modality preference based on their reliability and effectiveness.

\subsection{Proxy‑Aligned Co‑Evolution Mechanism}\label{evolution}
Static textual proxies provide a limited representation of the open-set negative space and are inherently sensitive to distributional shifts arising at test time. Adapting proxies with visual modality partially alleviates this issue but neglects complementary cross-modal cues that can enhance OOD discrimination. To overcome these limitations, we propose a \textit{Proxy-Aligned Co-Evolution} mechanism, in which textual and visual proxy caches are dynamically refined through bidirectional interactions. This co-evolution process enables the proxies to remain mutually aligned and better capture the semantics of both in-distribution and previously unseen OOD instances.

\paragraph{Textual Proxy Evolution.} 
Static textual proxies cannot faithfully track semantic shifts of test samples, so we evolve the textual negatives while guarding against error amplification. We first gate updates by a confidence margin around the adaptive threshold $\delta$ (as in AdaND~\cite{CaoZZLLZH25}; see \textbf{Appendix~A}). Concretely, samples with $\mathcal{S}_{\mathtt{CoEvo}}^{\mathrm{pre}}(x)>\delta+\gamma(1-\delta)$ are treated as ID, whereas those with $\mathcal{S}_{\mathtt{CoEvo}}^{\mathrm{pre}}(x)<\delta-\gamma(1-\delta)$ are treated as OOD; samples within the margin are excluded from updates to avoid uncertain decisions.

We adapt only the negative proxy queue $\mathbf{T}_n$ and keep the positive proxy queue $\mathbf{T}_p$ fixed to preserve stable anchors for ID semantics and prevent drift toward spurious OOD cues. Let $\mathcal{D}$ denote a corpus of $\ell_2$-normalized textual embeddings spanning a broad semantic vocabulary. Conditioned on the visual embedding $\mathbf{f}_{\mathrm{v}}$, we retrieve two candidate sets:
\begin{align}
\mathcal{N}_{\text{near}}(x) &= \operatorname{Top\text{-}N}_{e \in \mathcal{D}\setminus \mathbf{T}_n} \cos\!\big(\mathbf{f}_{\mathrm{v}}, e\big), \label{eq5} \\
\mathcal{N}_{\text{far}}(x)  &= \operatorname{Top\text{-}N}_{e \in \mathcal{D}\setminus \mathbf{T}_n} \big(-\cos\!\big(\mathbf{f}_{\mathrm{v}}, e\big)\big), \label{eq6}
\end{align}
with a deduplication constraint that discards candidates whose textual labels already appear in $\mathbf{T}_n$. We then update $\mathbf{T}_n$ by
\begin{equation}
\mathbf{T}_n \leftarrow
\begin{cases}
\big[\mathbf{T}_n;\ \mathrm{stack}\!\left(\mathcal{N}_{\text{near}}(x)\right)\big], & \text{if OOD}, \\
\big[\mathbf{T}_n;\ \mathrm{stack}\!\left(\mathcal{N}_{\text{far}}(x)\right)\big], & \text{if ID}.
\end{cases}
\end{equation}
Intuitively, inserting semantically \emph{near} negatives for predicted OOD samples tightens local open-set boundaries around the test sample, improving fine-grained separability from nearby ID proxies; inserting \emph{far} negatives for predicted ID samples broadens the coverage of the negative space, strengthening global discrimination against unseen classes. After each update, we recompute the textual score $\mathcal{S}_{\mathrm{T}}^{\mathrm{post}}(x)$ via Eq.~\eqref{eq1}, enabling progressive refinement of the textual proxies without modifying backbone weights.

\paragraph{Visual Proxy Evolution.}
Updating textual proxies alters the shared semantic manifold across modalities. Without corresponding adjustments on the visual side, this shift yields misaligned decision boundaries and degraded OOD discrimination. To preserve cross-modal consistency while improving the representational quality of visual proxies, we adopt an instance-adaptive strategy that refines the visual proxy cache at test time.

Given the updated negative textual proxy queue $\mathbf{T}_n$, we expand the negative visual proxy queue from $\mathbb{R}^{M\times L\times D}$ to $\mathbb{R}^{(M+N)\times L\times D}$ to accommodate the newly exposed OOD semantics, where $N$ matches the incremental textual negatives. Let $\{\mathbf{v}_k^p\}_{k=1}^{K}$ and $\{\mathbf{v}_m^n\}_{m=1}^{M+N}$ be the current visual proxies for ID and OOD (computed via Eq.~\eqref{eq2}). For $\mathbf{f}_{\mathrm{v}}$, 
we assign it to the most relevant proxy using soft similarity scores:
\begin{align}\label{eq89}
\mathbf{z}^p_k &= \mathrm{Softmax}\big(\cos(\mathbf{f}_{\mathrm{v}},\mathbf{v}_k^p)\big), 
\quad y_{\text{id}}=\arg\max_{k}\mathbf{z}^p_k,\\
\mathbf{z}^n_m &= \mathrm{Softmax}\big(\cos(\mathbf{f}_{\mathrm{v}},\mathbf{v}_m^n)\big),
\quad y_{\text{ood}}=\arg\max_{m}\mathbf{z}^n_m.
\end{align}

Based on the preliminary multi-modal score from Eq.~\eqref{eq4}, the sample $\mathbf{f}_{\mathrm{v}}$ is inserted into either the positive class-specific queue $\mathbf{V}_p^{y_{\text{id}}}$ (for ID samples) or the negative queue $\mathbf{V}_n^{y_{\text{ood}}}$ (for OOD samples), enabling the proxy cache to adaptively track evolving data distributions. To ensure reliability and prevent proxy drift, we exploit the entropy $\mathcal{H}(\mathbf{z})=-\sum_{i=1}\mathbf{z}_i \log \mathbf{z}_i$ as a confidence measure. A new sample is cached directly if space is available. Otherwise, it replaces the existing exemplar with the highest entropy (least confident) only if its own entropy is lower, ensuring that proxies are updated preferentially with high-confidence samples.

Following each update, we recompute the visual OOD detection score $\mathcal{S}_{\mathrm{V}}^{\mathrm{post}}(x)$ via Eq.~\eqref{eq3}, allowing the refined proxies to immediately influence detection outcomes. This instance-adaptive evolution maintains compact and representative visual proxies for ID classes, while preserving diverse and semantically rich proxies for OOD samples. Combined with textual proxy evolution, it ensures stable cross-modal alignment and enhances open-set discriminability under dynamic test-time conditions.

\paragraph{OOD Score Evolution.}
To ensure robust decision making during inference, we adapt the multi-modal OOD score to reflect the evolving proxy caches.

\textbf{Pre-evolution scoring.}
Following Eq.~\eqref{eq4}, the initial score is computed as
\begin{equation}
\mathcal{S}_{\mathtt{CoEvo}}^{\mathrm{pre}}(x)= \lambda\,\mathcal{S}_{\mathrm{T}}^{\mathrm{pre}}(x) + (1-\lambda)\,\mathcal{S}_{\mathrm{V}}^{\mathrm{pre}}(x) \quad \lambda\in[0.5,1).
\end{equation}
The higher textual weight realises a \emph{cold-start asymmetry}: stable semantic priors from pre-defined textual proxies are preferred while the visual proxies are still sparsely initialised.

\textbf{Post-evolution scoring.}
After the preliminary ID/OOD assignment and subsequent proxy evolution, we recompute the unimodal scores $\mathcal{S}_{\mathrm{T}}^{\text{post}}(x)$ and $\mathcal{S}_{\mathrm{V}}^{\text{post}}(x)$ and fuse them in a symmetric manner:
\begin{equation}\label{eq10}
\mathcal{S}_{\mathtt{CoEvo}}^{\mathrm{post}}(x) =
(1-\lambda)\,\mathcal{S}_{\mathrm{T}}^{\text{post}}(x) +
\lambda\,\mathcal{S}_{\mathrm{V}}^{\text{post}}(x).
\end{equation}
The weight flip is driven by two observations:
(i) \emph{Cold-start asymmetry:} as discussed above, and  
(ii) \emph{Post-adaptation reliability:} after evolution, the visual proxies, enriched with instance-specific samples, draw sharper local decision boundaries than their textual counterparts.

\textbf{Final decision.}
Unless otherwise specified, $\mathcal{S}_{\mathtt{CoEvo}}^{\mathrm{pre}}(x)$ is used solely for proxy updates, while $\mathcal{S}_{\mathtt{CoEvo}}^{\mathrm{post}}(x)$ is employed for the final ID/OOD decision (see Algorithm~\ref{alg:co-evo}).

\begin{algorithm}[t]
\caption{Cross-modal Proxy Evolving  (\texttt{CoEvo})}
\label{alg:co-evo}
\small
\begin{algorithmic}[1]
\REQUIRE ID label set $\mathcal{Y}_\mathrm{ID}$; text corpus $\mathcal{D}$ for negatives; test set $\mathcal{X}$; adaptive margin $\gamma$
\ENSURE Final ID/OOD predictions $\hat{\mathcal{Y}}$ for all $x \in \mathcal{X}$
\STATE Initialize positive textual queue $\mathbf{T}_p$ from $\mathcal{Y}_\mathrm{ID}$
\STATE Initialize negative textual proxy $\mathbf{T}_n$ from $\mathcal{D}$
\STATE Initialize visual proxy queues $\mathbf{V}_p, \mathbf{V}_n$ using $\mathbf{T}_p,\mathbf{T}_n$
\FOR{each sample $x \in \mathcal{X}$}
\STATE Compute textual score $\mathcal{S}_{\mathrm{T}}^{\mathrm{pre}}(x)$  (Eq.~\eqref{eq1})
\STATE Compute  visual score $\mathcal{S}_{\mathrm{V}}^{\mathrm{pre}}(x)$ (Eq.~\eqref{eq3})
\STATE Fuse  preliminary  multi-modal score $\mathcal{S}_{\mathtt{CoEvo}}^{\mathrm{pre}}(x)$  (Eq.~\eqref{eq4})
\STATE Compute the adaptive threshold $\delta$ 
\IF{$\mathcal{S}_{\mathtt{CoEvo}}^{\mathrm{pre}}(x) > \delta+\gamma(1-\delta)$} 
\STATE Predict preliminary label $\hat{y}_{\mathrm{id}}$
\STATE Retrieve far textual negatives $\mathcal{N}_{\mathrm{far}}(x)$ (Eq.~\eqref{eq6})
\STATE $\mathbf{T}_n \gets \text{Enqueue}(\mathbf{T}_n, \mathcal{N}_{\mathrm{far}}(x))$
\STATE $\mathbf{V}_p^{\hat{y}_{\mathrm{id}}} \gets \text{Enqueue}(\mathbf{V}_p^{\hat{y}_{\mathrm{id}}}, \mathbf{f}_\mathrm{v})$
\ELSIF{$\mathcal{S}_{\mathtt{CoEvo}}^{\mathrm{pre}}(x) < \delta(1-\gamma)$} 
\STATE Predict preliminary label $\hat{y}_{\mathrm{ood}}$
\STATE Retrieve near textual negatives $\mathcal{N}_{\mathrm{near}}(x)$ (Eq.~\eqref{eq5})
\STATE $\mathbf{T}_n \gets \text{Enqueue}(\mathbf{T}_n, \mathcal{N}_{\mathrm{near}}(x))$
\STATE $\mathbf{V}_n^{\hat{y}_{\mathrm{ood}}} \gets \text{Enqueue}(\mathbf{V}_n^{\hat{y}_{\mathrm{ood}}}, \mathbf{f}_\mathrm{v})$
    \ELSE
        \STATE Skip the proxy update for ambiguous sample
\ENDIF
    \STATE Update post-evolution scores $\mathcal{S}_{\mathrm{T}}^{\mathrm{post}}(x)$, $\mathcal{S}_{\mathrm{V}}^{\mathrm{post}}(x)$
    \STATE Fuse final score $\mathcal{S}_{\mathtt{CoEvo}}^{\mathrm{post}}(x)$ (Eq.~\eqref{eq10})
    \STATE Final decision $\hat{y}$ from $\mathcal{S}_{\mathtt{CoEvo}}^{\mathrm{post}}(x)$
\ENDFOR
\RETURN $\hat{\mathcal{Y}}$
\end{algorithmic}
\end{algorithm}

\section{Experiment}
\begin{table*}[tb] 
\centering
\resizebox{\linewidth}{!}{
\begin{tabular}{lcccccccc|cc}
\toprule
\multicolumn{11}{c}{OOD datasets}  \\
\multicolumn{1}{c}{\multirow{2}{*}{Methods}} 
& \multicolumn{2}{c}{iNaturalist} 
& \multicolumn{2}{c}{Sun} 
& \multicolumn{2}{c}{Places} 
& \multicolumn{2}{c}{Textures} 
& \multicolumn{2}{c}{Average} 
\\ \cmidrule(lr){2-3} \cmidrule(lr){4-5} \cmidrule(lr){6-7} \cmidrule(lr){8-9} \cmidrule(lr){10-11}
 & AUROC$\uparrow$ & FPR95$\downarrow$
 & AUROC$\uparrow$ & FPR95$\downarrow$
 & AUROC$\uparrow$ & FPR95$\downarrow$
 & AUROC$\uparrow$ & FPR95$\downarrow$ 
 & AUROC$\uparrow$ & FPR95$\downarrow$  \\
 \midrule
MSP \cite{hendrycks2016baseline}    &    87.44 & 58.36 & 79.73 & 73.72 & 79.67 & 74.41 & 79.69 & 71.93 & 81.63 & 69.61   \\
Energy \cite{liu2020energy}  &    95.33 & 26.12 & 92.66 & 35.97 & 91.41 & 39.87 & 86.76 & 57.61 & 91.54 & 39.89 \\
GradNorm \cite{huang2021importance} &   72.56 & 81.50 & 72.86 & 82.00 & 73.70 & 80.41 & 70.26 & 79.36 & 72.35 & 80.82 \\
NPOS \cite{tao2023non}  &    96.19 & 16.58 & 90.44 & 43.77 & 89.44 & 45.27 & 88.80 & 46.12 & 91.22 & 37.93\\
ZOC \cite{esmaeilpour2022zero}               & 86.09 & 87.30 & 81.20 & 81.51 & 83.39 & 73.06 & 76.46 & 98.90 & 81.79 & 85.19 \\
CLIPN \cite{wang2023clipn}            & 95.27 & 23.94 & 93.93 & 26.17 & 92.28 & 33.45 & 90.93 & 40.83 & 93.10 & 31.10 \\
LoCoOp \cite{miyai2024locoop} & 96.86 & 16.05 & 95.07 & 23.44 & 91.98 & 32.87 & 90.19 & 42.28 & 93.52 & 28.66 \\
LAPT \cite{zhang2024lapt}  & 99.63 & 1.16 & 96.01 & 19.12 & 92.01 & 33.01 & 91.06 & 40.32 & 94.68 & 23.40 \\
NegPrompt \cite{li2024learning} & 98.73 & 6.32 & 95.55 & 22.89 & 93.34 & 27.60 & 91.60 & 35.21 & 94.81 & 23.01 \\
 \midrule
Energy \cite{liu2020energy}           & 85.09 & 81.08 & 84.24 & 79.02 & 83.38 & 75.08 & 65.56 & 93.65 & 79.57 & 82.21 \\
MCM \cite{ming2022delving} & 94.59 & 32.20 & 92.25 & 38.80 & 90.31 & 46.20 & 86.12 & 58.50 & 90.82 & 43.93 \\
NegLabel \cite{jiang2024negative} &99.49&1.91&95.49& 20.53 & 91.64 & 35.59 & 90.22 & 43.56 & 94.21 & 25.40 \\
CSP \cite{Chen0X24} & 99.60 & 1.54 & 96.66 & 13.66& 92.90& 29.32& 93.86 &25.52& 95.76& 17.51\\
AdaNeg \cite{ZhangZ24} & 99.71 & 0.59 & 97.44 & 9.50 & 94.55 & 34.34 & 94.93 & 31.27 & 96.66 & 18.92 \\
\midrule
\textbf{\texttt{CoEvo}}$_\mathrm{NegLabel}$ & 99.81 & 0.53 & \textbf{98.68} & \textbf{4.42} & \textbf{95.80} & \textbf{23.51} & \textbf{97.48} & \textbf{12.42} & \textbf{97.95} & \textbf{10.22} \\
\textbf{\texttt{CoEvo}}$_\mathrm{CSP}$ & \textbf{99.82} & \textbf{0.46} & 98.61 & {4.68} & {95.58} & {25.83} & {97.38} & {12.78} & {97.85} & {10.94} \\
\bottomrule
\end{tabular}}\vspace{-1mm}
\caption{
Comparison of OOD detection performance on ImageNet-1K. 
All methods utilize a CLIP ViT-B/16 encoder.}\label{tab1}\vspace{-2mm}
\end{table*}

\subsection{Experimental Setup}
\paragraph{Datasets.} 
Following prior work \cite{ming2022delving,jiang2024negative}, we conduct extensive experiments on the ImageNet-1K benchmark~\cite{deng2009imagenet}, where the large-scale ImageNet-1K dataset is used as ID source. Four commonly adopted datasets, iNaturalist~\cite{van2018inaturalist}, SUN~\cite{xiao2010sun}, Places~\cite{zhou2017places}, and Textures~\cite{cimpoi2014describing}, are employed as OOD test sets. To further evaluate generalization under varying OOD difficulty, we adopt both Near-OOD (SSB-hard~\cite{vaze2021open}, NINCO~\cite{bitterwolf2023ninco}) and Far-OOD (iNaturalist~\cite{van2018inaturalist}, Textures~\cite{cimpoi2014describing}, OpenImage-O~\cite{wang2022vim}) settings, as defined by the OpenOOD benchmark~\cite{zhang2023openood,yang2022openood}, where ImageNet serves as the shared ID dataset. In addition, we examine performance under imbalanced conditions between ID and OOD samples to assess the robustness of our approach in realistic deployment settings.

%%%%%%%%%%%%可以放附录%%%%%%%%%%%%%%%
\paragraph{Evaluation Criteria.} 
Following prior work~\cite{ming2022delving}, we adopt three standard metrics for evaluating OOD detection performance:(1) \textbf{FPR95:} the false positive rate of OOD samples when the true positive rate (TPR) of ID samples is at $95\%$; (2) \textbf{AUROC:} the area under the receiver operating characteristic curve, which measures the overall separability between ID and OOD samples; and (3) \textbf{ID ACC:} classification accuracy for ID samples.

\paragraph{Implementation Details.} We employ the ViT-B/16 visual encoder pretrained by CLIP~\cite{radford2021learning} as our backbone. The key hyperparameters are set as follows:
the visual proxy queue length is $L=10$; 
the temperature $\tau=0.01$ in Eq.~\eqref{eq1} and \eqref{eq3};
the balancing weight $\lambda=0.8$ in Eqs.~\eqref{eq4} and \eqref{eq10};
$N=5$ in Eq.~\eqref{eq5}; and $\beta = 5.5$ in Eq.~\eqref{eq2}.  The batch size is fixed at $128$. The lexical database and corresponding negative mining  strategy for the textual proxy cache are derived from two recent baselines: NegLabel~\cite{jiang2024negative} and CSP~\cite{Chen0X24}, which initialize negative textual proxies with $10\text{K}$ and $9{,}493$ OOD class names, respectively. Following NegLabel, we adopt the  prompt template ``The nice \texttt{cls}'' to encode class names. All experiments are conducted on a single NVIDIA RTX 3090 GPU.

\begin{table}[t!]
\centering
\resizebox{\linewidth}{!}{
\begin{tabular}{lccccc}
\toprule
\multirow{2}{*}{\textbf{Method}} 
& \multicolumn{2}{c}{\textbf{FPR95} $\downarrow$} 
& \multicolumn{2}{c}{\textbf{AUROC} $\uparrow$} 
& \multicolumn{1}{c}{\textbf{ACC} $\uparrow$} \\
\cmidrule(lr){2-3} \cmidrule(lr){4-5}
& Near-OOD & Far-OOD & Near-OOD & Far-OOD & ID \\
\midrule
GEN             & -- & -- & 78.97 & 90.98 & 81.59 \\
AugMix + ReAct  & -- & -- & 79.94 & 93.70 & 77.63 \\
RMDS            & -- & -- & 80.09 & 92.60 & 81.14  \\
SCALE           & -- & -- & 81.36 & 96.53 &  76.18 \\
AugMix + ASH    & -- & -- & 82.16 & 96.05 & 77.63 \\
LAPT            & 58.94 & 24.86 & 82.63 & 94.26 & 67.86 \\
\midrule
MCM             & 79.02 & 68.54 & 60.11 & 84.77 & 66.28 \\
NegLabel        & 69.45 & 23.73 & 75.18 & 94.85 & 66.82\\
CSP             &  73.14 &21.52 & 74.88 & 95.87 & 67.35\\  
AdaNeg          & 67.51 & 17.31 & \textbf{76.70} & 96.43 & 67.13 \\
\midrule
\textbf{\texttt{CoEvo}}$_\mathrm{NegLabel}$ & \textbf{64.64} & {15.24} & {75.37} & {96.50} & {66.83} \\
\textbf{\texttt{CoEvo}}$_\mathrm{CSP}$ & {66.88} & \textbf{14.47} & {74.65} & \textbf{96.70} & \textbf{67.36} \\
\bottomrule
\end{tabular}}\vspace{-1mm}
\caption{
Zero-shot OOD detection results on the OpenOOD benchmark,
where ImageNet-1K is adopted as the ID dataset. 
}\label{tab2}\vspace{-2mm}
\end{table}

\subsection{Main Results}
\paragraph{Evaluation on ImageNet benchmark.} As shown in Tab.~\ref{tab1}, our method consistently outperforms existing methods, including both training-based methods~\cite{hendrycks2016baseline,liu2020energy,huang2021importance,tao2023non,esmaeilpour2022zero,wang2023clipn,miyai2024locoop,zhang2024lapt,li2024learning} and training-free methods~\cite{liu2020energy,ming2022delving,jiang2024negative,Chen0X24,ZhangZ24}. Specifically, CoEvo$_\mathrm{NegLabel}$ achieves an average FPR95 of $10.22\%$ and AUROC of $97.95\%$, outperforming the most competitive baseline by margins of $45.98\%$ in FPR95 and $1.33\%$ in AUROC, respectively.

\paragraph{Evaluation on OpenOOD benchmark.} As illustrated in Tab.~\ref{tab2}, our method achieves competitive performance under both Near-OOD and Far-OOD settings. Note that training-based baselines leverage the full ImageNet training set. Under Near-OOD conditions, \textbf{CoEvo}$_\mathrm{CSP}$ obtains an average FPR95 of $66.88\%$ and an AUROC of $74.65\%$, slightly underperforming AdaNeg~\cite{ZhangZ24} in AUROC, indicating marginally reduced sensitivity to fine-grained OOD discrimination. Conversely, in Far-OOD scenarios, our method achieves a substantially lower average FPR95 of $14.47\%$ and a high AUROC of $96.70\%$. Furthermore, our approach improves ID classification performance, achieving an average ID ACC of $67.36\%$, surpassing all competing training-free methods. These results collectively validate the effectiveness and robustness of our proposed framework across diverse OOD settings.

\begin{table}[t!]
    \centering
    \resizebox{0.7\linewidth}{!}{
    \begin{tabular}{cccc}
        \toprule
        \multicolumn{2}{c}{\textbf{Proxy Evolution}} & \multicolumn{2}{c}{\textbf{Average}} \\
        \cmidrule(lr){1-2} \cmidrule(lr){3-4}
        Textual & Visual & FPR95~$\downarrow$ & AUROC~$\uparrow$ \\
        \midrule
        & & 24.97 & 94.56 \\
        \checkmark & & 21.77 & 95.38 \\
        & \checkmark & 17.41 & 96.99 \\
        \checkmark & \checkmark & \textbf{10.22} & \textbf{97.95} \\
        \bottomrule
    \end{tabular}}\vspace{-2mm}
    \caption{Ablation study of proxy evolution mechanism on ImageNet-1K, evaluated across four standard OOD datasets.}
    \label{Tab:3}
    \vspace{-4mm}
\end{table}

\subsection{Analyses and Discussions}
\paragraph{Analysis of Proxy Evolution.}
We perform an ablation study to quantify the contribution of each component in the proposed proxy-aligned co-evolution mechanism (Tab.\ref{Tab:3}). The baseline, NegLabel without any evolution step, achieves an average FPR95 of $24.97\%$ and an AUROC of $94.56\%$. Activating {textual evolution} alone lowers the FPR95 to $21.77\%$, revealing that dynamically updating text proxies during test-time evolution enhances semantic alignment. Enabling {visual evolution} alone produces a greater gain, confirming that adapting visual proxies to the test distribution mitigates feature-space shifts. Combining both textual and visual evolution yields the best performance, an average FPR95 of $10.22\%$ and an AUROC of $97.95\%$, highlighting the complementary strengths of cross-modal co-evolution for robust OOD detection.

\begin{figure}[t!]
    \centering
    \includegraphics[width=0.45\textwidth]{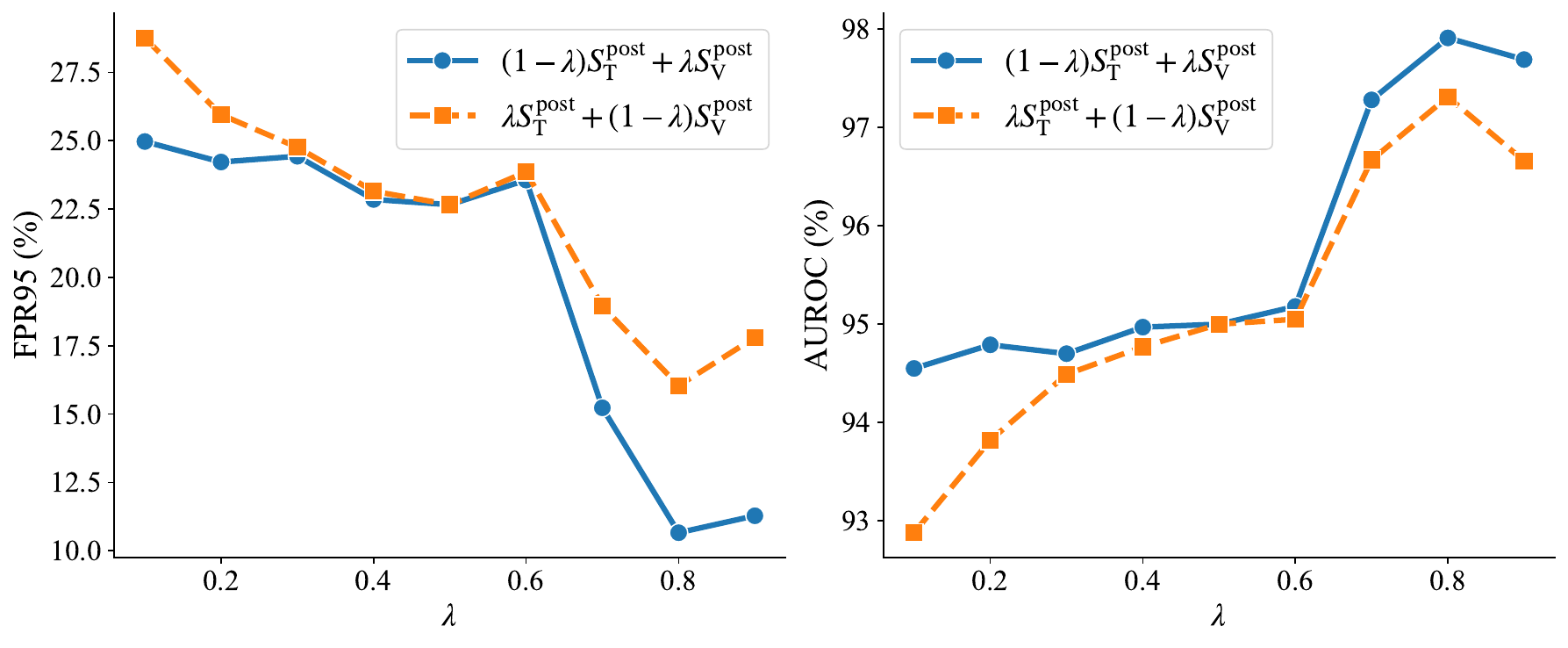}\vspace{-2mm}
    \caption{Sensitivity to the fusion weight $\lambda$. Results are reported on the ImageNet-1K benchmark. 
    % Dashed lines denote the fixed score $\mathcal{S}_{\mathtt{CoEvo}}^{\mathrm{pre}}$, while solid lines denote the evolved score $\mathcal{S}_{\mathtt{CoEvo}}^{\mathrm{post}}$.
    }
    \label{fig:lam}\vspace{-1mm}
\end{figure}

\paragraph{Analysis of Score Evolution.} 
In Fig.\ref{fig:lam}, we compare the proposed post-evolution fusion in Eq.\eqref{eq10} (solid line) against a no-flip variant that retains the pre-evolution weighting scheme, i.e., $\lambda\mathcal{S}_{\mathrm{T}}^{\text{post}}(x)+(1-\lambda)\mathcal{S}_{\mathrm{V}}^{\text{post}}(x)$ (dashed line). Low-$\lambda$ regime ($0.1$–$0.4$): Eq.~\eqref{eq4} allocates higher weight to visual scores; however, early in evolution the visual cache is sparsely populated, producing noisy estimates and unstable OOD separation. The no-flip strategy inherits this bias and overemphasizes unreliable visual evidence, while the flipped rule shifts preference toward the more stable textual cues, achieving better performance. High-$\lambda$ regime ($0.6$–$0.9$): As evolution progresses, the visual proxies become richer and more discriminative due to the accumulation of diverse samples. The flipped rule adaptively assigns greater weight to these refined visual scores, surpassing the no-flip variant, with the performance gap peaking around $\lambda=0.8$. These results demonstrate that fixed weight retention fails to adapt to the evolving reliability of modalities, whereas our score evolution mechanism dynamically aligns fusion weights with proxy quality, consistently improving OOD detection performance.

\paragraph{Analysis of Hyper-parameter $\lambda$.}
We conduct an ablation study on the fusion weight $\lambda$ in Eq.~\eqref{eq10}. As shown in Fig.~\ref{fig:lam}, performance on ImageNet-1K first increases with $\lambda$, peaking at $\lambda=0.8$, and then slightly degrades as $\lambda$ approaches $1.0$. This trend indicates that a moderate emphasis on the visual score, combined with complementary textual cues, yields the best performance. When $\lambda$ is too small, the model over-relies on textual proxies, which are generally coarse and less adaptive to instance-specific features. The optimal setting of $\lambda=0.8$ thus achieves a trade-off between the adaptability of visual proxy and the semantic stability of textual proxy.

\begin{figure}[t!]
    \centering
    \includegraphics[width=0.5\textwidth]{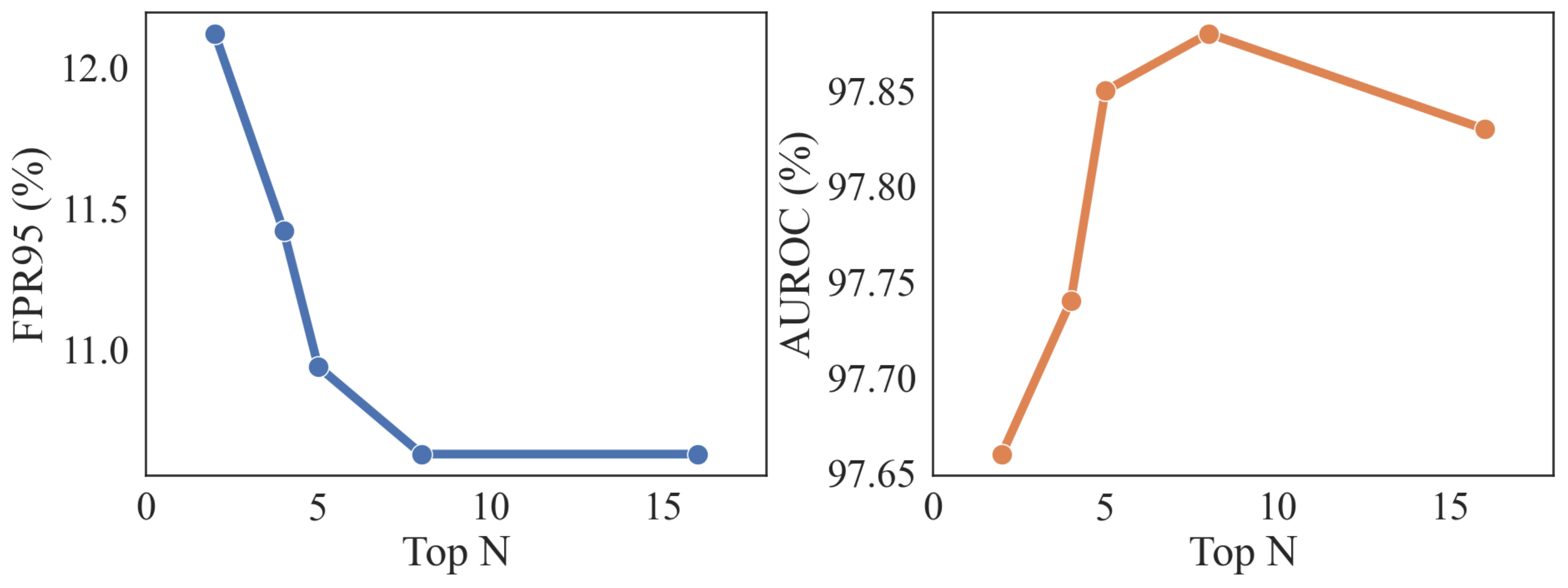}\vspace{-2mm}
    \caption{Sensitivity analysis of the hyperparameter $\operatorname{Top}\text{-}N$, evaluated on the ImageNet-1K benchmark.}
    \label{fig:topn}\vspace{-1mm}
\end{figure}

\paragraph{Impact of $\operatorname{Top\text{-}N}$ in Eq.~\eqref{eq5}.}
We investigate how varying the retrieval parameter $N$ in Eq.~\eqref{eq5} influences the evolution of textual proxies in $\mathtt{CoEvo}_\mathrm{CSP}$ on ImageNet-1K (Fig.~\ref{fig:topn}). With a small $N$, both $\mathcal{N}_{\text{near}}$ and $\mathcal{N}_{\text{far}}$ exhibit limited diversity, resulting in a sparsely populated negative queue $\mathbf{T}_n$ and weakened OOD discrimination. Increasing $N$ initially improves performance by injecting semantically richer negatives, thereby refining textual adaptation. However, beyond a task-dependent threshold, the gains saturate and may even decline due to two factors despite the deduplication constraint: (i) the marginal similarity gap between successive candidates diminishes, introducing redundancy rather than novel information; and (ii) a larger $N$ increases the likelihood of enqueuing weakly aligned or noisy candidates. Empirically, we set $N=5$, which offers a favorable trade-off between semantic coverage, prediction stability, and computational efficiency.

\begin{table}[t!]
    \centering
    \resizebox{0.95\linewidth}{!}{
    \begin{tabular}{l|cccccc}
    \toprule
    \textbf{ID:OOD Ratio} & 1:100 & 1:10 & 1:1 & 10:1 & 100:1 \\
    \midrule
    NegLabel        & 23.00 & 20.50  & 21.55 & 25.92 & 19.69 \\
    CSP             & 20.00 & 18.00 & 16.78 & 19.50 & 17.95 \\
    AdaNeg          & 28.00 & 14.00  & 8.01  & 10.08 & 17.40  \\
    \textbf{\texttt{CoEvo}}$_\mathrm{NegLabel}$  & 17.00 &  \textbf{6.70} & \textbf{5.27} & \textbf{5.76} & \textbf{14.77} \\ 
    \textbf{\texttt{CoEvo}}$_\mathrm{CSP}$ & \textbf{14.00} & {7.50}  & {5.58} & {6.15} & {15.38}\\
    \bottomrule
    \end{tabular}}\vspace{-1mm}
    \caption{FPR95 ($\downarrow$) under different ID:OOD mixture ratios on ImageNet-1K (ID) and SUN (OOD).}\vspace{-2mm}
    \label{tab:4}
\end{table}

\paragraph{Robustness to Data Imbalance.}
We further evaluate the robustness of our $\mathtt{CoEvo}$ under varying ID-OOD data imbalance scenarios. Five experimental settings are constructed using ImageNet-1K as ID and SUN as OOD data. \textbf{(i) Low-ID regimes:} For ratios of $1{:}100$, $1{:}10$, and $1{:}1$, we randomly sample $10\,\text{K}$ SUN images as OOD data, paired with $100$, $1\,\text{K}$, and $10\,\text{K}$ ImageNet samples, respectively.  
\textbf{(ii) High-ID regimes:} For ratios of $10{:}1$ and $100{:}1$, we fix $10\,\text{K}$ ImageNet samples as ID data, combined with $1\,\text{K}$ and $100$ SUN images as OOD data. As shown in Tab.~\ref{tab:4}, our method consistently outperforms all baselines across all imbalance ratios. Notably, it maintains strong performance even under extreme imbalance (e.g., $100{:}1$ with only $100$ OOD samples), demonstrating robustness to real-world data distribution shifts.

% \paragraph{Extend Analysis and Ablations.}
% Further analyses and ablations on the queue length $L$, adaptive threshold $\delta$, and other experimental results are provided in \textbf{Appendix~B}.
\section{Conclusion}

In this work, we introduced \texttt{CoEvo}, a test-time zero-shot OOD detection framework that enables bidirectional, sample-conditioned adaptation across visual and textual modalities. \texttt{CoEvo} maintains modality-specific proxy caches and iteratively refines them through a proxy-aligned co-evolution mechanism, dynamically realigning cross-modal similarity under distribution shifts without updating the backbone parameters. Furthermore, a multi-modal score evolution strategy fuses dual-modal evidence to produce calibrated OOD scores. Extensive experiments on standard benchmarks validate the effectiveness of our approach, demonstrating consistent improvements over negative-label baselines.  Beyond the proposed framework, this work highlights the importance of dynamic, cross-modal adaptation for robust open-world recognition, paving the way for future research on scalable, label-free OOD detection.

% \section*{Acknowledgments}
% This work was supported by the Shenzhen-Hong Kong-Macao Science and Technology Plan Project (Category C Project) under the Shenzhen Municipal Science and Technology Innovation Commission (Project No. SGDX20230821092359002) and a grant under the Collaborative Research with World-leading Research Groups scheme of The Hong Kong Polytechnic University (project no. G-SACF). This research was also supported by the Guangdong Natural Science Funds for Distinguished Young Scholars (Grant No. 2023B1515020097), the National Research Foundation Singapore under its AI Singapore Programme (AISG Award Nos.: AISG3-GV-2023-011 and AISG4-TC-2025-018-SGKR), the Singapore Ministry of Education AcRF Tier 1 Grant (Grant No. MSS25C004), and the Lee Kong Chian Fellowships.

% \bibliographystyle{aaai}  
% \bibliography{reference.bib} %!!!

% \clearpage
% \input{Checklist/ReproducibilityChecklist}
\clearpage
\section{Appendix}

This appendix provides additional technical details, experimental results, theoretical analysis, and discussions to complement the main paper. Specifically:

\begin{itemize}
    \item \textbf{Appendix~A} details the adaptive thresholding strategy used for confidence-based proxy updates, including the definition and role of the threshold $\delta$ and margin $\gamma$.
    \item \textbf{Appendix~B} presents extended experimental results of \texttt{CoEvo}, including ablation studies, sensitivity analyses, and complexity analyses.
    \item \textbf{Appendix~C} provides a more comprehensive survey of related work, particularly test-time adaptation and vision-language models.
    % \item \textbf{Appendix~D} offers a theoretical perspective on the limitations of existing negative-label methods, and analyzes why proxy-aligned co-evolution improves cross-modal alignment.
    \item \textbf{Appendix~D} discusses the limitations of our proposed \texttt{CoEvo} framework.
\end{itemize}

\section{Appendix A: Adaptive Threshold $\delta$}
Threshold-based decision rules are widely used in zero-shot OOD detection. Prior methods, such as AdaNeg~\cite{ZhangZ24}, employ a fixed threshold to separate positive from negative samples. However, this assumption implicitly presumes that the distribution of OOD scores is stable across different test environments. In practice, ID and OOD samples encountered at test time may vary significantly in both scale and separation margin, making a fixed threshold $\delta$ suboptimal and potentially leading to unreliable OOD decisions.

To address this issue, we adopt a \emph{data-driven adaptive threshold} that adjusts dynamically to the score distribution of the incoming test data. Inspired by AdaND~\cite{CaoZZLLZH25}, we estimate $\delta$ by minimizing the intra-class variance of the multi-modal OOD scores.

\subsection{Adaptive Threshold Estimation}
Let each test sample $x_i$ be associated with a preliminary multi-modal score $S_i=\mathcal{S}_{\mathtt{CoEvo}}^{\mathrm{pre}}(x_i)$. For a candidate threshold $\delta$, we define the sets of predicted ID and OOD samples:
\begin{equation}
\begin{aligned}
N_{\text{id}}(\delta) &= \bigl|\{\,i: S_i>\delta\}\bigr|,\quad
\mu_{\text{id}}(\delta) = \frac{1}{N_{\text{id}}(\delta)}\sum_{j: S_j>\delta} S_j,\\
N_{\text{ood}}(\delta) &= \bigl|\{\,i: S_i\le\delta\}\bigr|,\quad
\mu_{\text{ood}}(\delta) = \frac{1}{N_{\text{ood}}(\delta)}\sum_{j: S_j\le\delta} S_j.
\end{aligned}
\end{equation}
The optimal threshold $\delta^*$ is obtained by minimizing the overall intra-class variance of the scores:
\begin{equation}
\begin{aligned}
\delta^* =\arg\min_{\delta}\Biggl\{
&\frac{1}{N_{\text{id}}(\delta)}
    \sum_{i: S_i>\delta}\bigl(S_i-\mu_{\text{id}}(\delta)\bigr)^2 \\
&+
\frac{1}{N_{\text{ood}}(\delta)}\sum_{i: S_i\le\delta}\bigl(S_i-\mu_{\text{ood}}(\delta)\bigr)^2
\Biggr\}.
\end{aligned}
\label{eq:adaptive_beta}
\end{equation}
This formulation adaptively chooses the threshold that yields the most compact ID and OOD score clusters, thereby improving their separability and reducing sensitivity to distribution shifts. The estimation of $\delta^*$ can be performed incrementally over a sliding window of test samples, ensuring online adaptability without labeled OOD data.

\subsection{Confidence-Aware Proxy Updates}
While $\delta^*$ provides an adaptive decision boundary, samples located near this threshold remain highly ambiguous and may degrade the quality of the evolving proxy caches if inserted indiscriminately. To mitigate this issue, we introduce a \emph{confidence margin} parameter $\gamma\in[0,1]$ to filter out low-confidence samples. The refined decision rule becomes:
\begin{align}
\text{Negative:} & \quad S_i<\delta^*(1-\gamma)\notag\\
\text{Positive:} & \quad S_i\ge\delta^*+\gamma(1-\delta^*),
\label{eq:confidence_gap}
\end{align}
and any sample satisfying
\begin{equation}
\delta^*(1-\gamma)\le S_i<\delta^*+\gamma(1-\delta^*)
\end{equation}
is considered uncertain and excluded from proxy updates.
This selective caching strategy ensures that only high-confidence samples contribute to the proxy-aligned co-evolution mechanism, preventing the propagation of noisy or unstable evidence and improving the reliability of test-time adaptation.

\section{Appendix B: Additional Results}

% \begin{table}[t!]
% \centering
% \caption{Zero-shot OOD detection results of \textbf{CoEvo}$_\mathrm{CSP}$ on the OpenOOD benchmark, where ImageNet-1k is used as the ID dataset.}\label{tab:detailedopenood}
% \resizebox{0.9\linewidth}{!}{
% \begin{tabular}{llcc}
% \toprule
% \textbf{OOD Type} & \textbf{Dataset} & \textbf{FPR95 $\downarrow$} & \textbf{AUROC $\uparrow$} \\
% \midrule
% \multirow{3}{*}{Near-OOD} 
% & SSB-hard       &  72.24 & 72.01  \\
% & NINCO          &  61.52 &77.30   \\
% & \textbf{Mean}  &  66.88 & 74.65  \\
%  \midrule     
% \multirow{4}{*}{Far-OOD} 
% & iNaturalist   & 0.57 &  99.82 \\
% & Textures      & 12.78 &  97.11 \\
% & OpenImage-O   & 30.07 & 93.17 \\
% & \textbf{Mean} & 14.47 & 96.70 \\
% \bottomrule
% \end{tabular}}
% \vspace{-0.2cm}
% \end{table}

\begin{table*}[t!] 
	\centering
	
    \resizebox{\linewidth}{!}{  
	\begin{tabular}{lcccccccccc}
		\toprule
		\multicolumn{1}{c}{\multirow{2}{*}{\textbf{Backbone}}} &
\multicolumn{2}{c}{iNaturalist} &
\multicolumn{2}{c}{Sun} &
\multicolumn{2}{c}{Places} &
\multicolumn{2}{c}{Textures} &
\multicolumn{2}{c}{Average} \\
\cmidrule(lr){2-3} \cmidrule(lr){4-5} \cmidrule(lr){6-7} \cmidrule(lr){8-9} \cmidrule(lr){10-11}
& AUROC $\uparrow$ & FPR95 $\downarrow$ 
& AUROC $\uparrow$ & FPR95 $\downarrow$ 
& AUROC $\uparrow$ & FPR95 $\downarrow$ 
& AUROC $\uparrow$ & FPR95 $\downarrow$ 
& AUROC $\uparrow$ & FPR95 $\downarrow$ \\
		\midrule
		ResNet-50 & 99.7  & 0.93 &  98.05  &5.91 &  95.06  &37.23 & 96.68 & 14.16 & 97.37 & 14.56\\
		ViT-B/32 & 99.82 & 0.57  & 98.51 & 4.3  &  95.34 & 28.63 & 97.63 & 10.44 & 97.82  &10.98\\
		ViT-B/16 & {99.82} & {0.46} & 98.61 & {4.68} & {95.58} & {25.83} & {97.38} & {12.78} & {97.85} & {10.94} \\
		\bottomrule
	\end{tabular}
    }
    \caption{Zero-shot OOD detection results of the proposed \texttt{CoEvo} across different CLIP networks on the ImageNet-1K benchmark.} \label{tab:5}
\end{table*}

\subsection{Anaylsis of CLIP Networks} 
To assess the robustness of \texttt{CoEvo} with respect to the underlying vision-language backbone, we evaluate the OOD detection performance of \texttt{CoEvo}$_\mathrm{CSP}$ across several CLIP architectures. As shown in Tab.~\ref{tab:5}, ViT-B/16 delivers the best overall results, achieving an average AUROC of $97.85\%$ and FPR95 of $10.94\%$. Notably, ViT-B/32 attains the lowest FPR95 ($10.44\%$) on the Textures dataset, suggesting that larger patch sizes may benefit fine-grained texture discrimination. These observations indicate that \texttt{CoEvo} maintains strong OOD detection capability across different CLIP variants, demonstrating its adaptability to diverse backbone choices without requiring architecture-specific tuning.

\begin{figure}[t!]
    \centering
    \includegraphics[width=0.45\textwidth]{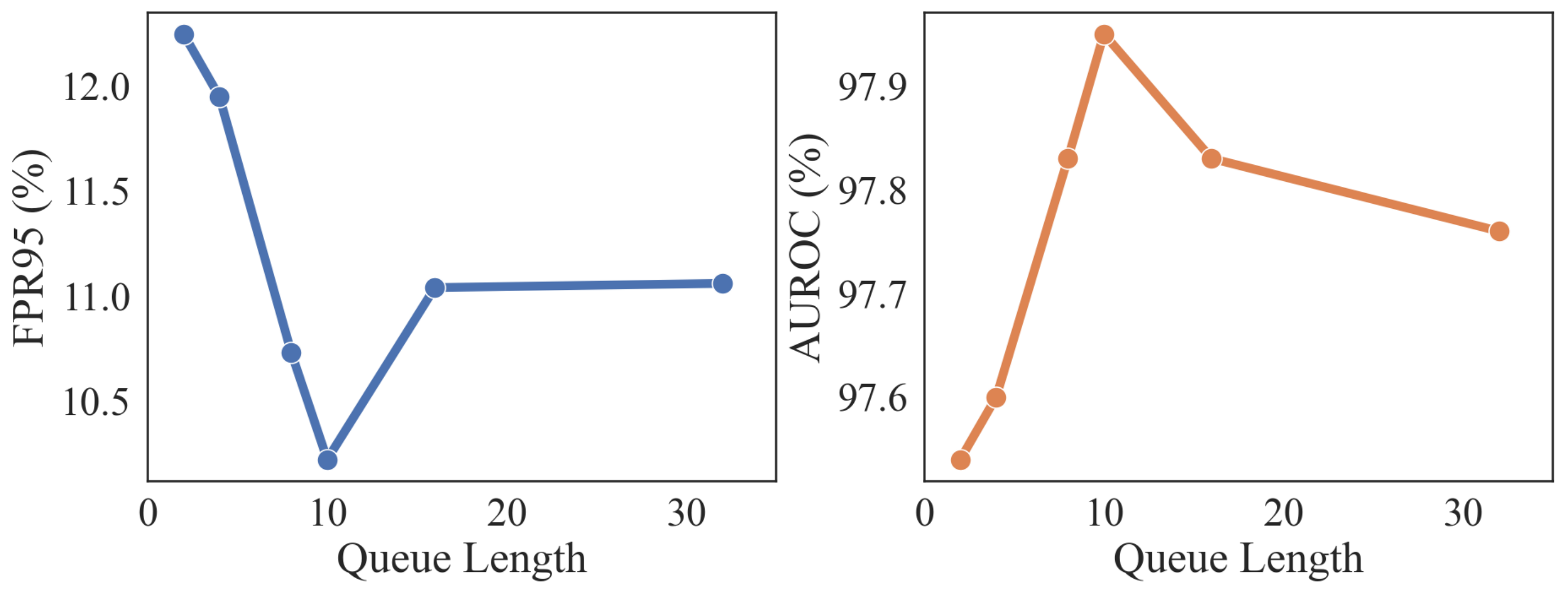}
    \caption{Analyses on the hyper-parameter of queue length $L$, where results are reported on ImageNet-1K benchmark with \texttt{CoEvo}$_\mathrm{CSP}$.}
    \label{fig:qlen}
\end{figure}

\subsection{Impact of Queue Length $L$}
The queue length $L$ controls the number of visual samples retained per class in the positive/negative proxy caches, directly influencing the diversity and freshness of visual proxies during co-evolution. We evaluate the sensitivity of CoEvo$_\mathrm{CSP}$ to varying values of $L$ on the ImageNet-1k benchmark (Fig.~\ref{fig:qlen}). As $L$ increases from small values, AUROC improves steadily, indicating that a richer memory enhances alignment between test samples and stored proxies. However, beyond a moderate length, performance gains saturate and eventually decline slightly, likely due to the accumulation of stale or less relevant samples, which introduces redundancy and noise. These findings highlight the importance of maintaining a compact yet informative proxy cache, and we set $L=10$ in all experiments to achieve a favorable balance between diversity and stability.

\begin{figure}[t!]
    \centering
    \includegraphics[width=0.45\textwidth]{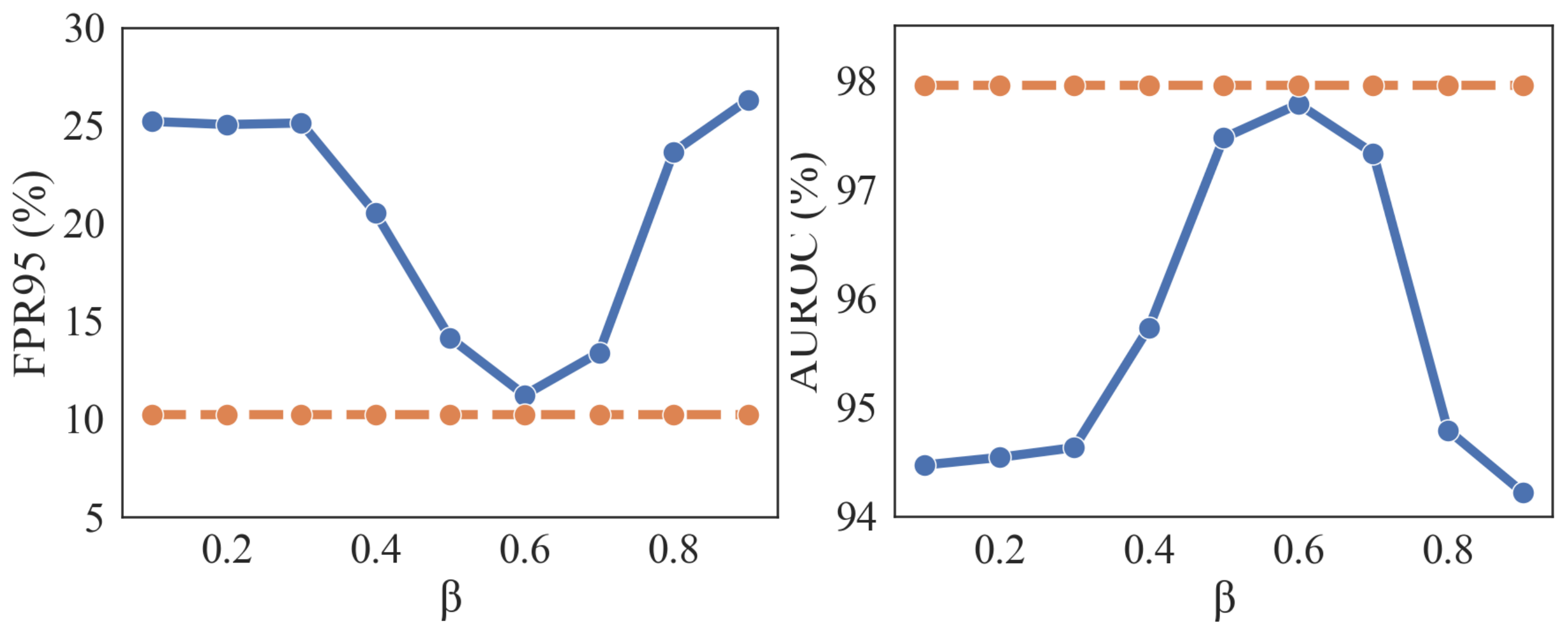}
    \caption{Impact of the threshold $\delta$ on OOD detection performance, evaluated on the ImageNet-1K benchmark with \texttt{CoEvo}$_\mathrm{NegLabel}$. 
    Solid lines correspond to fixed thresholds $\delta$, while dashed lines represent the proposed adaptive threshold.
    }
    \label{fig:thres}
\end{figure}

\subsection{Adaptive Threshold vs. Fixed Threshold}
To assess the reliability and robustness of the proposed adaptive threshold, we compare its performance with fixed thresholds $\delta \in \{0.1,0.2,\dots,0.9\}$ on the ImageNet-1K benchmark. The results are summarized in Fig.~\ref{fig:thres}, where solid lines denote fixed thresholds and dashed lines represent the adaptive threshold. Our analysis reveals that the adaptive threshold consistently outperforms all fixed settings in terms of both FPR95 and AUROC, demonstrating its effectiveness in dynamically adjusting to the underlying score distribution. Beyond performance improvements, the adaptive threshold also provides a crucial practical benefit: it removes the need for exhaustive hyperparameter tuning across different ID datasets, significantly enhancing the applicability of \texttt{CoEvo} in real-world, open-world scenarios. Consequently, we employ the adaptive threshold in all reported experiments.

\begin{figure}[tb!]
    \centering
    \includegraphics[width=0.45\textwidth]{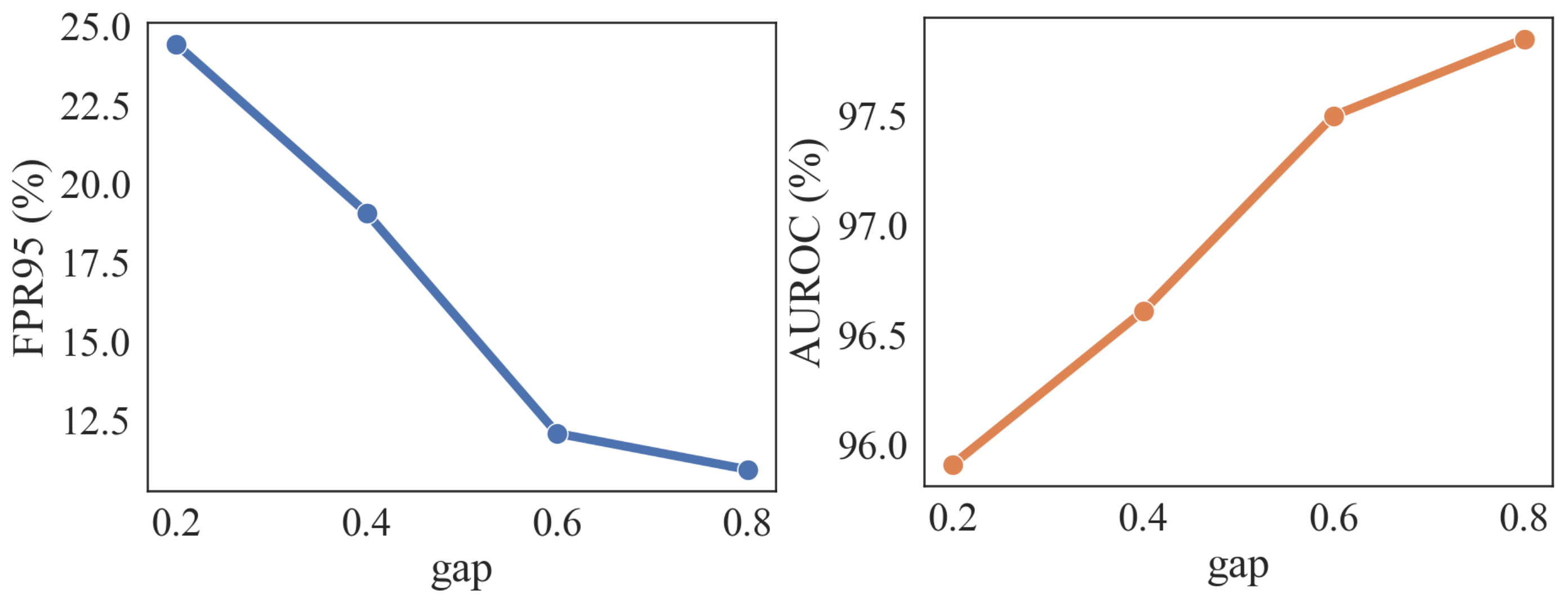}
    \caption{Sensitivity analysis of the confidence margin $\gamma$ on the ImageNet-1K benchmark using \texttt{CoEvo}$_\mathrm{NegLabel}$.}
    \label{fig:gap}
\end{figure}

\subsection{Sensitivity to the Margin $\gamma$}
The margin $\gamma$ regulates the confidence region used for updating the proxy cache. When $\gamma=0$, all samples are cached, including those with low ID/OOD separation confidence, potentially introducing noisy or misleading visual proxies. Increasing $\gamma$ progressively filters out ambiguous samples near the adaptive threshold, improving the purity of stored proxies and stabilizing the co-evolution process. As shown in Fig.~\ref{fig:gap}, larger $\gamma$ values consistently yield higher AUROC and lower FPR95 on ImageNet-1K, confirming that selective caching improves OOD discrimination. We also observe that gains saturate beyond a moderate margin, suggesting a trade-off between proxy quality and the number of samples retained for adaptation.

\begin{table}[t!]
    \centering
    \resizebox{\linewidth}{!}{
    \begin{tabular}{lccc}
        \toprule
        \textbf{Method} & \textbf{FPS} & \textbf{Param.} & \textbf{FPR95} $\downarrow$ \\
        \midrule
      NegLabel~\cite{jiang2024negative} & 962 & -& 25.25 \\
          CSP~\cite{Chen0X24} & 967 & -&21.33 \\
          AdaNeg~\cite{ZhangZ24} & 504 & -&17.10 \\
          $\texttt{CoEvo}_\mathrm{NegLabel}$ & 408 & -&10.22 \\
          $\texttt{CoEvo}_\mathrm{CSP}$ & 420 & -&10.94 \\
        \bottomrule
    \end{tabular}}
    \caption{Efficiency analysis of \texttt{CoEvo} and competing methods on the ImageNet-1K benchmark. ``FPS'' denotes inference speed (batch size = 128), and ``Param.'' is the number of learnable parameters. All results are obtained on an NVIDIA RTX3090 GPU.}
    \label{Tab:cost}
\end{table}

\subsection{Complexity Analyses} 
We report the inference speed and performance of \texttt{CoEvo} and representative baselines in Tab.~\ref{Tab:cost}. All evaluations are conducted on the ImageNet-1K benchmark using an NVIDIA RTX 3090 GPU with a batch size of 128. Notably, \texttt{CoEvo} introduces no additional learnable parameters and requires no model retraining, making it lightweight and deployment-friendly. Despite its parameter-free nature, \texttt{CoEvo} achieves substantial improvements in detection accuracy, reducing FPR95 by $15.03\%$ and $10.39\%$ compared to NegLabel and CSP, respectively.
In terms of efficiency, \texttt{CoEvo} exhibits slightly reduced FPS relative to baseline methods such as NegLabel and CSP, due to the online maintenance and update of dual-modal proxy caches during inference. Nevertheless, this trade-off is justified by the significant performance gains, especially under large-scale benchmarks. Compared to AdaNeg, which also performs test-time visual adaptation, \texttt{CoEvo} offers superior detection (FPR95 of $10.22\%$ vs. $17.10\%$) at a comparable inference speed (408 vs. 504 FPS), highlighting its efficiency–accuracy balance. Overall, \texttt{CoEvo} provides a compelling solution for real-world deployment scenarios where robustness to unseen inputs is critical and limited adaptation cost is acceptable.

\begin{table}[t!]
    \centering
    \resizebox{\linewidth}{!}{
    \begin{tabular}{lccccc}
    \toprule
    \textbf{Method} & \textbf{90} & \textbf{900} & \textbf{9K} & \textbf{45K} & \textbf{90K} \\
    \midrule
    NegLabel & 24.00 &	17.00	&22.30&	21.45	&21.61 \\
    CSP & 8.00 & 12.00&	17.32	&16.80&	17.18\\
    AdaNeg   & 10.00 &12.00 &	8.82	&10.53	&10.06 \\
   $\texttt{CoEvo}_\mathrm{NegLabel}$ &4.00 &5.80 & 6.44 & 4.99 & 4.62\\
    $\texttt{CoEvo}_\mathrm{CSP}$ & 4.00&	6.40&	6.72	&5.32	&4.73\\
    \bottomrule
    \end{tabular}}
    \caption{Impact of test sample quantity on FPR95 ($\downarrow$). We evaluate each method with increasing numbers of test images (90 to 90K), where samples are randomly drawn from ImageNet (ID) and SUN (OOD) with a fixed ID:OOD ratio of 5:4.}
    \label{tab:test_num}
\end{table}

\subsection{Impact of Test Set Size} 
Tab.~\ref{tab:test_num} presents the performance (FPR95) of each method under varying numbers of test samples, ranging from 90 to 90K. All settings use a fixed ID:OOD ratio of 5:4, with samples drawn from ImageNet (ID) and SUN (OOD). This experiment simulates real-world streaming conditions and examines the scalability and stability of different methods under limited or abundant test data. 

\texttt{CoEvo} consistently outperforms all baselines across all sample sizes. Notably, even with as few as 90 or 900 test samples, both $\texttt{CoEvo}_\mathrm{NegLabel}$ and $\texttt{CoEvo}_\mathrm{CSP}$ achieve substantially lower FPR95 compared to methods like NegLabel ($24.00\%$ vs. $4.00\%$ at 90 samples) and CSP ($8.00\%$ vs. $4.00\%$). This demonstrates the strong initial discriminative capability of our proxy-aligned co-evolution mechanism, which can rapidly adapt from limited unlabeled data without requiring prior tuning or auxiliary training signals.
As the number of test samples increases (from 900 to 90K), the cached proxy sets in \texttt{CoEvo} become progressively enriched, leading to further improvements in FPR95 (e.g., from $6.44\%$ to $4.62\%$ for $\texttt{CoEvo}_\mathrm{NegLabel}$). In contrast, most baseline methods exhibit stagnation or fluctuating performance, indicating a limited ability to incorporate evolving data distributions.

Interestingly, with only 90 test samples, the OOD detection task becomes underdetermined relative to the label space size (e.g., 1000 ID classes in ImageNet-1K), simplifying the binary decision boundary and partially explaining the exceptionally low FPR95 in this regime. Nonetheless, the consistent advantage of \texttt{CoEvo} across all data scales highlights its robustness and test-time adaptability under both low-data and large-scale deployment scenarios.

\section{Appendix C: Related Work}

\subsection{Test-time Adaptation}
Test-time adaptation (TTA)~\citep{wang2021tent, liang2023ttasurvey, niu2022efficient, boudiaf2022parameter, prabhudesai2023diffusion, lee2024entropy, gui2024atta} aims to enhance a model’s generalization to target distributions by continuously adapting it using a stream of unlabeled test samples. Since access to source-domain data is typically restricted during inference, recent works have proposed various adaptation strategies to mitigate performance degradation.

Early TTA approaches~\citep{wang2021tent, niu2022efficient} leverage self-supervised objectives, such as entropy minimization, while others rely on test-time batch normalization statistics~\citep{schneider2020improving} to correct distributional shifts. More recently, growing interest has emerged in adapting large-scale vision-language models (VLMs) under test-time constraints~\citep{shu2022test, feng2023diverse, karmanov2024efficient, ma2024swapprompt, zhao2024testtime, yoon2024ctpt}. For instance, TPT~\citep{shu2022test} and DiffTPT~\citep{feng2023diverse} optimize test-time prompts using entropy-based feedback from individual samples. TDA~\citep{karmanov2024efficient} introduces a training-free dynamic adapter for efficient VLM adaptation, and DMN~\cite{zhang2024dual} employs a dynamic memory module to retain contextual information across test samples.

Despite these advancements, most existing TTA methods focus on single-modality adaptation, limiting their capacity to fully capture task-specific semantics in out-of-distribution (OOD) scenarios. In contrast, \texttt{CoEvo} is designed to co-evolve both textual and visual proxy representations during test time, enabling bidirectional cross-modal adaptation. This joint evolution progressively refines multi-modal semantics and improves OOD awareness under distribution shifts.

\subsection{Pre-trained Vision-Language Models}
Pre-trained vision-language models (VLMs) such as CLIP~\citep{radford2021learning}, ALIGN~\citep{jia2021scaling}, and GroupViT~\citep{xu2022groupvit} consist of paired image and text encoders, jointly trained on hundreds of millions of image–text pairs via self-supervised contrastive learning~\citep{chen2020simple}. During inference, VLMs embed both input images and textual queries into a shared feature space, and perform classification by computing the similarity between the resulting embeddings.

Thanks to large-scale training and contrastive objectives~\citep{LuLK23,LuWLLK24,LuZLZK25}, VLMs exhibit strong generalization to a wide range of downstream tasks~\cite{FuLY0SSD21,0002WWYJLL25,Wu0YHFJYL24,WuZLCJ25,WuLZZBZR24,WangMZWFXX24,fu2024cross,fu2024objectrelator}, including image classification and retrieval, particularly in zero-shot settings. In this work, we tackle the problem of zero-shot out-of-distribution (OOD) detection, where the model must operate solely during test time, without access to training samples or ground-truth labels from the target domain. This setting is critical for real-world deployment scenarios, where prior annotation is infeasible, and robustness to unseen environments is essential.

\section{Appendix D: Limitations}

Despite the strong empirical performance and robust test-time adaptability demonstrated by \texttt{CoEvo}, several limitations remain.

First, our framework assumes the availability of a large-scale, semantically diverse textual corpus to enable negative proxy mining. In resource-constrained scenarios where such corpora are unavailable or domain-specific language is underrepresented, the quality of proxy evolution may degrade.

Second, although \texttt{CoEvo} operates in a zero-shot setting without requiring labeled OOD samples, it still relies on a well-trained vision-language backbone (e.g., CLIP). In domains with significant distribution shifts from the pretraining corpus (e.g., medical imaging or remote sensing), the pre-trained embedding space may inadequately capture task-relevant semantics, thereby limiting co-evolution effectiveness.

Third, the current proxy update mechanism is heuristic and confidence-driven, which may be suboptimal in highly noisy or adversarial test-time environments. A more theoretically grounded or uncertainty-aware update policy could further improve robustness.

Lastly, the proposed method incurs moderate computational overhead due to the maintenance and updating of proxy queues during inference. Although this cost is amortized over the test stream, future extensions could explore more efficient memory management or lightweight co-evolution strategies.

We leave addressing these limitations as promising directions for future work.

% \clearpage

\bibliography{reference.bib} %!!!

\end{document}